\def\year{2020}\relax
\documentclass[letterpaper]{article} % DO NOT CHANGE THIS
\usepackage{aaai20}  % DO NOT CHANGE THIS
\usepackage{times}  % DO NOT CHANGE THIS
\usepackage{helvet} % DO NOT CHANGE THIS
\usepackage{courier}  % DO NOT CHANGE THIS
\usepackage[hyphens]{url}  % DO NOT CHANGE THIS
\usepackage{graphicx} % DO NOT CHANGE THIS
\usepackage{enumitem}
\usepackage{graphicx}  % DO NOT CHANGE THIS
\title{A Spoken Dialogue System for Spatial Question Answering in a Physical Blocks World}
\author{
\textbf{Georgiy Platonov, Benjamin Kane, Aaron Gindi, Lenhart K. Schubert} \\ Department of Computer Science, \\ University of Rochester, \\ Rochester, NY, USA \\ gplatono@cs.rochester.edu, bkane2@u.rochester.edu, agindi@u.rochester.edu, schubert@cs.rochester.edu 
}

\begin{document}

\maketitle

\begin{abstract}
The blocks world is a classic toy domain that has long been used to build and test spatial reasoning systems. Despite its relative simplicity, tackling this domain in its full complexity requires the agent to exhibit a rich set of functional capabilities, ranging from vision to natural language understanding. There is currently a resurgence of interest in solving problems in such limited domains using modern techniques. In this work we tackle spatial question answering in a holistic way, using a vision system, speech input and output mediated by an animated avatar, a dialogue system that robustly interprets spatial queries, and a constraint solver that derives answers based on 3-D spatial modeling. The contributions of this work include a semantic parser that maps spatial questions into logical forms consistent with a general approach to meaning representation, a dialog manager based on a schema representation, and a constraint solver for spatial questions that provides answers in agreement with human perception. These and other components are integrated into a multi-modal human-computer interaction pipeline.
\end{abstract}

\section{Introduction}
The past 10 or 20 years have seen rapid progress in many areas of AI technology. However, despite impressive advances in specific, narrow tasks, such as object recognition, natural language parsing and machine translation using RNNs and word and sentence embeddings, game playing, etc., there is still a shortage of multimodal interactive systems capable of performing high-level tasks requiring understanding and reasoning. Because of the complexity of most real-life tasks, the blocks world domain provides an ideal experimental setting for developing prototypes with such capabilities. 

Interest in the blocks world as a domain for AI research goes back as far as the 1970s, with Winograd's thesis \cite{winograd1972understanding} being one of the earliest studies, utilizing a virtual environment along with text-based interaction. Recently there has been a resurgence of interest in solving problems in such limited domains using modern techniques. Despite its relative simplicity, the blocks world domain motivates implementation of diverse capabilities in a virtual interactive agent aware of physical blocks on a table, including visual scene analysis, spatial reasoning, planning, learning of new concepts, dialogue management and voice interaction, and more. In this work, we describe an end-to-end system that integrates several such components in order to perform a simple task of spatial question answering about block configurations.

As will be seen in the following sections, our system is distinctive in that (1) it is an end-to-end system using computer vision and spoken dialogue with an on-screen virtual human; (2) it did not require a large training corpus, only a modest development corpus using naturally posed spatial questions by a few participants; (3) the development of the semantic parser took only a few weeks -- no more, we infer, than semantic parsers developed for NN-based spatial QA studies; yet it produces logical forms consistent with a comprehensive approach to English logical form, rather than producing specialized procedures, like some of the most nearly comparable systems; (4) the system models the perceived scene in 3D graphics; (5) the constraint-based spatial relation models are realistic, rather than artificially simplified, as for example in many CLEVR-based studies; (6) scenes and the dialogues are real, rather than synthetic; and (7) quite good results are achieved with symbolic and soft constraint-solving methods alone. 

\section{Related Work}
Early studies featuring the blocks world include \cite{winograd1972understanding} and \cite{fahlman1974planning}, both of which relied on a simulated environment. The latter was focused on construction planning, rather than user interaction, and as such incorporated extensive reasoning about geometric consistency and structural stability, more than descriptive aspects of the block configurations. Modern efforts in blocks world spatial language, in a similar spirit as ours, include work by Perera et al.~\cite{perera2018situated,perera2018building}. Their studies are focused on learning spatial concepts in the blocks world (such as staircases, towers, etc.) based on verbally-conveyed structural constraints, e.g., \textit{``The height is at most 3''}, as well as explicit examples and counterexamples, given by user, that help the system to zero in on the correct set of constraints that define the structure class. Though aimed at goals quite different from ours, their framework, like ours, includes physical blocks on a table, Kinect cameras that capture the scene, speech communication, and a commitment to deriving utterance meanings that fit into a comprehensive approach to meaning representation.

The work by Bisk et al.~\cite{bisk2018learning} 
%involves spatial relations since it focuses on 
is concerned with learning
%spatial relations in the blocks world using a large dataset of synthetically generated scenes. 
how to move individual blocks in a simulated 3D world in accord with typed instructions. These typically involve spatial relations, e.g., \textit{``Move McDonald's so it’s just to the right (not touching) the Twitter block"}. The system learns to transduce such commands into block displacements (based on numerous variants collected through crowdsourcing).
% The training and testing also make use of block coordinates and 3D scene representations, as provided by the model.% I JUST DON'T KNOW IF THIS IS TRUE
How thoroughly the system needs to understand the linguistic inputs in order to pick the right action is hard to assess, since it is not designed to explicitly describe spatial relations among blocks, or answer questions about them.

The CLEVR dataset \cite{johnson2017clevr} and its modified versions, such as \cite{liu2019clevr},
% MODIFIED BY LS:
can be viewed as a response to this kind of semantic obliquity in prior work, since it lays out an explicit spatial question answering challenge. This has
inspired a flurry of projects on visual reasoning. % Citations are now prefixed to discussions below
% ADDED BY LS:
One noteworthy example, \cite{mao2019neuro}, achieves near-perfect scores on the CLEVR questions; however, the images (of blocks, cylinders and
% The scores are actually superhuman, but I decided not to mention it, because it's due to the artificiality  of the data
spheres) and the questions used for training and testing are synthetically generated, and ground-truth answers are based on greatly simplified models of the available relations: \textit{behind} means further away in the (fixed) depth direction, regardless of lateral position or intervening objects, \textit{left} means any amount laterally to the left, regardless of depth or intervening objects, etc. Human judgments are far more subtle, but it would be hard to create realistic (non-synthetic) training corpora of adequate size for NN-based approaches. \cite{kottur2019clevr} adds interesting dialogue aspects to CLEVR -- information-providing statements alternating with information-seeking questions, along with coreference capabilities, but the training and testing data remain synthetic. % END OF ADDITION 
% I THINK WE SHOULD OMIT THIS-- IT DOESN'T HAVE MUCH TO DO WITH SPATIAL RELATION QA  
%\cite{asai2018photo} also relies on the CLEVR-derived dataset, and describes a hybrid neuro-symbolic approach to blocks world planning. The approach employs variational autoencoders trained on image pairs that represent the consecutive states of the world before and after a certain action was performed. The autoencoders learn latent representations for the states of the world and the set of possible state-to-state transitions. The latent vector representations extracted in this way can be used in conjunction with any classical propositional planner based on searching a state space. 
%
So while
the domain and scope of CLEVR-based studies are similar to ours in some respects, e.g., in referring to colors and demonstrating counting-related reasoning, they differ significantly in other respects, especially their reliance on synthetic data, their two-dimensional (image-based) rather than three-dimensional modeling of the table-top world, their unrealistic and very limited ground-truth models of spatial relations, and their use of domain-specific procedural formalisms for linguistic semantics. 

% The work \cite{Zhang2019LearningTC} explores the task of collaborative structure building in a 2D blocks world context. They use a listener-speaker architecture, where both roles are filled by neural networks. Initially, a structure example is generated, using a probabilistic generative CFG. The ``speaker'' network receives the generated structure as an image and generates a sequence of symbols from a fixed finite alphabet (these symbols store learned latent representations and, as such are hard to interpret); these symbols are passed to the ``listener'' network which generates the action probabilities for the actions available in the CFG.

Since our work relies on explicit modeling of a perceived physical blocks world, we should also mention some approaches to such modeling. In blocks world planning following the STRIPS paradigm \cite{fikes1971strips}, models used for logical planning were generally purely qualitative, using relations such as \textit{on, in,} and \textit{next-to}. Dimensions, orientations and distances were typically treated as lower-level properties not directly relevant to reasoning. Generalizing from such work and from developments in qualitative temporal reasoning, many studies of spatial reasoning tended to take more of a topological rather than geometric view of spatial entities,  
treating them as sets of regions and introducing certain relations over the elements of these sets. These relations usually included contiguity, meronymy (part-whole relationship), direction, ordering, etc. %Let's skip this...
% As in the work on plan reasoning, the primitives tended to align with those in cognitive linguistics models (e.g., contiguity), which is natural in that descriptions of spatial configurations expressed in language are typically qualitative. 
The region connection calculus epitomizes such methods ~\cite{dhelim2016stlf,cohn2008qualitative,chen2015survey}.

%Since the structure of time is similar to that of one-dimensional space, there were attempts to apply this approach to spatial knowledge representation and reasoning. There is even a possibility of generalizing that approach to higher dimensions by taking projections of the objects onto coordinate axes and then applying these relations to these one-dimensional projections \cite{guesgen1989spatial}. However, significant numbers of assumptions and simplifications need to be made in order for this approach to work. Unfortunately, while this interval approach is very efficient for representing one-dimensional worlds, it is very hard to adapt it to realistic physical world situations since the two- and three-dimensional worlds are much richer in structure and content. Freksa \cite{freksa1991qualitative} also considers the interval approach in connection with his aquarium world and gives some methodological insights into the problem of generalizing temporal relations to cover higher-dimensional cases.

However, studies of how people actually judge spatial relations show that no crisp, qualitative models can do justice to those judgments. In a previous study
\cite{platonov2018computational}, we explored computational models for prepositions using imagistic modeling, akin to the current work (here, by ``imagistic'' we mean direct visual reconstruction of the perceived scene, a \textit{mental image}, that our system creates and operates on). Spatial relations were computed by evaluating geometric and non-geometric relations and properties of 3D models of objects in a scene, such as their bounding boxes, distances, etc. The paper considered two domains, a blocks world and a ``room world'' filled with common items, such as furniture, books, appliances, etc. Another study, similar in spirit, was \cite{bigelow2015need}, which applied an imagistic approach to a story understanding task. This study used Blender to create a three-dimensional scene and reason about the relative configuration and visibility of objects in the scene. At the core of spatial relation models was the notion of an acceptance area, i.e., a prismatic region corresponding to a particular spatial relation. 
%Each acceptance area is a prism with a rectangular base, representing the set of points for which the given spatial relation holds. 
For example, a requirement for truth of $A$ \textit{on} $B$ was that $A$ should lie in the acceptance area located directly above $B$. Probabilistic reasoning was supported by using values from 0 to 1 reflecting the volumetric proportion of the relatum falling into the relevant acceptance area.

Another example of an imagistic reasoning system was implemented as part of the planning system for the robot Ripley \cite{roy2004mental}. Ripley used two different representations of space. One  represented the three-dimensional structures in Ripley's surroundings, while the second represented a two-dimensional view of the world, coming from Ripley's cameras.
%The first model provided the 3D imagistic framework that served to represent the long-term global state of the world. 
The 3D model contained representations of Ripley's body, of the human operator (who communicated with Ripley in natural language), and Ripley's workspace. Several times per second, the model maintenance system updated the global 3D model, using the 2D view to check the existence and current status of objects in the world. If some new object was detected in the field of view of the robot, or some object stored in the global model was no longer present at its old position, or some properties of the object differed from those previously perceived, the global state model was updated to reflect the new state of the world. Prepositional spatial relations were modeled by Gaussian distributions in terms of Regier's spatial features \cite{regier1996human}: the off-horizontal tilt of the centroid-to-centroid line, the distance between the two most proximal points of the objects, and the off-horizontal tilt of the line joining these two points. No quantitative measure of success was reported, but Ripley was able to understand and carry out requests such as \textit{``Pickup the large green cup to the left of the blue plate"}.
It is no surprise that significant amounts of research on locative expressions and spatial relations are produced in the modern robotics. Using natural language is the most efficient way to issue a command to a robot, and since they have to operate in the physical world, understanding the way humans describe space if crucial. Current approaches to grounding natural language commands in general, and spatial commands in particular, are frequently based on probabilistic graphical models (PGM) such as \textit{Generalized Grounding Graphs} ($G^3$) ~\cite{tellex2011understanding} and \textit{Distributed Correspondence Graphs} (DCG) \cite{howard2014natural} and their modifications \cite{broad2016towards,paul2016efficient,boteanu2016model,chung2015performance,paul2018efficient}. 

In recent years, attempts have been made to use statistical learning models, especially deep neural networks, to learn spatial relations. Noteworthy examples are \cite{bisk2018learning} (already mentioned) and \cite{chang2014learning}. The latter study inverted the learning problem, in a sense; the task was not to learn how to describe object relationships, but rather to automatically generate a scene based on a textual description. Another recent study in this area is \cite{yu2017sentence}, wherein spatial relation models are used to locate and identify similar objects in several video streams. We should also mention \cite{collell2017acquiring}, which applies deep neural networks to learning spatial templates for triplets of form (relatum, relation, referent). The latter work does this in an implicit setting, that is, it uses relations that indirectly suggests certain spatial configurations, e.g., \textit{(person, rides, horse)}. Their model is capable not only of learning a spatial template for specific arguments but also of generalizing that template to previously unseen objects; e.g., it can infer the template for \textit{(person, rides, elephant)}. These approaches, however, rely on the analysis of 2D images rather than attempting to model relations in an explicitly represented 3D world.

%All of the methods discussed in this section have serious common drawbacks. First of all, they all were developed as a means to reason about space in a formal manner. While they do succeed at this task to a certain degree, the resulting formalisms are very unnatural, in the sense that they do not directly correspond to the spatial relations used by humans. Therefore, while they are useful, for example, in robot navigation, they are not directly applicable to modeling the prepositions and understanding natural language instructions. Another problem is that such models typically oversimplify the model of the world. For example, positional calculi approximate the world by a plane. This simplification leads to the loss of the important details about the configurations of objects. Based on concepts that are too idealized and abstract, such methods cannot capture the rich conceptualizations of the physical world. Because of these limitations, the domain where those methods are efficient is large-scale navigation, but not object manipulation.

\section{Task Description}
Our goal is dialogue-based question answering about spatial configurations of blocks on a table. The system is designed to answer straightforward questions such as {\it ``Which blocks are touching some red block?'', ``Is the X block clear?'', ``Where is the Y block?'',} etc.\ (where X and Y are unique block labels). The task of describing and answering questions about block configurations in a dialogue setting, while relatively simple, serves two purposes. First, it is function-complete in the sense that it requires most of the components needed for physical blocks world problem solving, such as dialogue management, audio-visual input-output, etc., to be in place. Once this pipeline is complete, one can proceed to add more advanced functionalities. Second, it sets the stage for our longer-term goal of building a collaborative blocks world agent, capable of interactively learning new structural concepts and building examples of them, relying on natural language communication with the user. This task requires a spatial reasoning component, operating on the ordinary kinds of spatial language used by people. 
%Thus, building a spatial QA system is a way to develop and test such a spatial language component.

We have collected an initial sample set of about 500 question-answer pairs to evaluate the breadth of the concepts and phrasings people use in describing or asking about spatial configurations. Three volunteers with no experience with the system participated in the data collection. This set was filtered to remove incoherent/contrived occurrences and the remaining filtered subset was used for guiding the development.

We roughly classify questions in one of six categories: identification, confirmation, existential, counting, descriptive, and attribute-inquiry questions. {\it ``Which block is...?''/``What is...''} is a general template for the questions of the first category, {\it ``Is X {\small IN RELATION} Y to Z?''} is a template for the second, etc. The support for basic spatial relations is implemented for individual blocks as arguments. We are working on the implementation for structures and regions as well, in order to be able to answer questions like {\it ``Is the bottom block in the tallest stack red?'', ``Which blocks are near the front edge of the table?''}, etc.

\section{Blocks World System Overview}
\begin{figure}[ht]
    \centering
    \includegraphics[width=0.9\columnwidth]{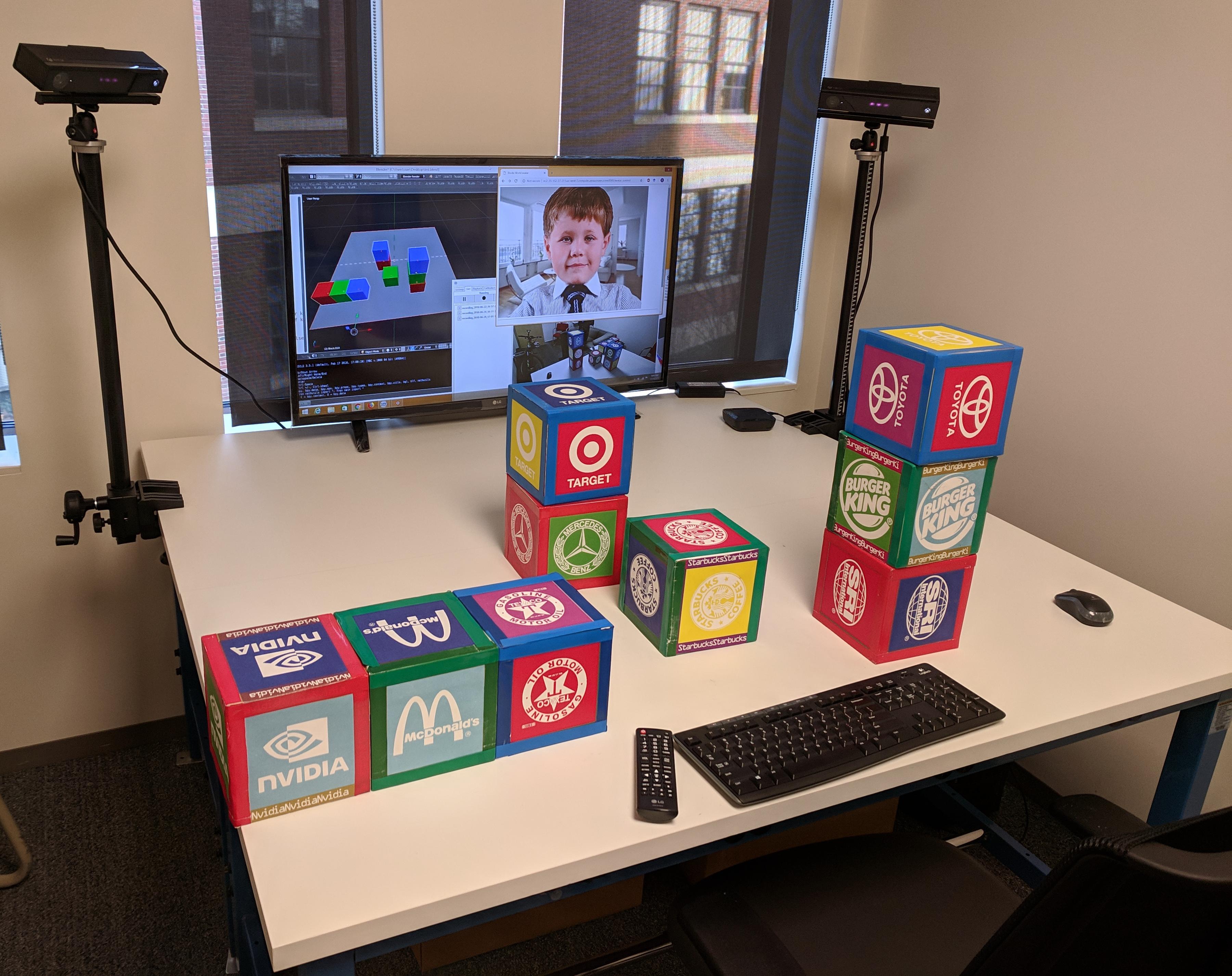}
    \caption{The blocks world apparatus setup.}
    \label{fig:bw_setup}
    %\Description{The blocks world setup}
\end{figure}

Our blocks world system consists of two components: the physical apparatus and a blocks world dialogue system\footnote{A site for the implementation of all but the Kinect blocks detector will be supplied.}.
% The implementation of most of the components (apart from the Kinect blocks detector) can be found at https://github.com/gplatono/spatialQA}.
The physical apparatus (see Fig. \ref{fig:bw_setup})
%is the embodiment of the blocks world domain and 
is comprised of a square table surface, approximately 1.5m x 1.5m in size, several cubical blocks with 0.15m sides, two Microsoft Kinect sensors mounted at the back end of the table to track the state of the world, and a display for user interaction. The blocks are marked with corporate logos, such as McDonald's, Toyota, Texaco, etc., which serve as block names and allow the user and the system to uniquely identify and refer to individual blocks. The blocks are also color-coded as either red, green, or blue, using the colored stripes running along the edges of the blocks (see the Figure).

The following example interaction between the user and the system demonstrates the kind of back-and-forth exchange our system allows for (based on the block properties):
\begin{itemize}[noitemsep]
    \item[] David: Would you like to ask me a spatial question?
    \item[] User: OK. Is the Texaco block to the right of the McDonalds block?
    \item[] David: Yes, it is.
    \item[] User: What blocks are above the Mercedes block?
    \item[] David: Only the Target block.
    \item[] User: What is the highest block?
    \item[] David: The Toyota block.
    \item[] User: How many blocks are between a red block and a blue block?
    \item[] David: Two blocks: The McDonalds block and the Burger King block.
    \item[] User: What blocks are near the Starbucks block?
    \item[] David: The Texaco, Mercedes, Target, and SRI blocks.
\end{itemize}

\begin{figure}[ht]
    \centering
    \includegraphics[width=0.9\columnwidth]{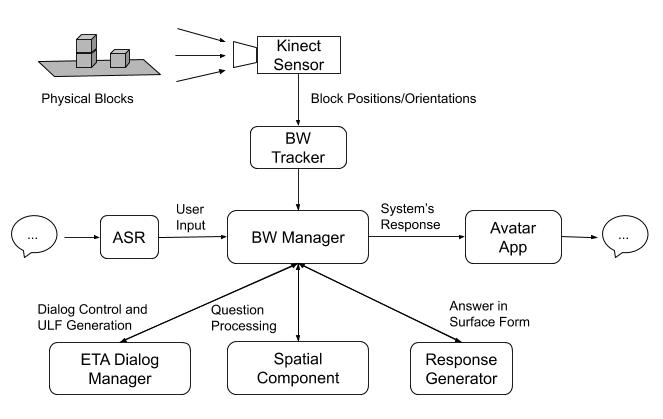}
    \caption{The blocks world dialogue pipeline. The arrows indicate the direction of interaction between the modules.}
    \label{fig:bw_pipeline}
    %\Description{The blocks world Dialogue Pipeline}
\end{figure}

The architecture of the software component is shown in Fig.~\ref{fig:bw_pipeline}. The system uses audio-visual input and output. The block detection and tracking module periodically reads the input from the Kinect cameras and updates the block positioning information. Based on the information from the block tracking module, the physical block arrangement is modeled as a 3D scene in Blender. All the spatial processing is performed on that model. The automatic speech recognition (ASR) module, based on the Google Cloud Speech-To-Text API, is responsible for generating the transcripts of user utterances. For communicating back to user, we employ the interactive avatar character, David, developed by SitePal. The avatar is capable of vocalizing the text and displaying facial expressions, making the flow of conversation more natural than with textual I/O. The spatial component module together with the constraint solver is responsible for analyzing the block configuration with respect to the conditions implicit in the user's utterance. The Eta dialogue manager is responsible for {\it unscoped logical form} (ULF) generation (see subsections on ULF and the dialog manager below) and controlling the dialogue flow and transition between phases, such as greeting, ending the session, etc. Finally, the blocks world manager is the unifying component, controlling the rest of the system and facilitating the message passing and synchronization between the modules.

A typical round of interaction starts with the user asking a spatial question. The blocks world manager obtains and preprocesses the transcript of user's speech, and then sends it to Eta. Eta generates the ULF representation of the question and sends it back to the blocks world manager. Having received the ULF string, the blocks world manager performs query processing, which includes two stages. First, the ULF string is parsed into an internal query frame, with slots corresponding to the main predicate in the question, arguments and modifiers of the predicate (e.g., negation, etc.). The advantage of this query frame is that it provides a more canonical representation of the question than the surface form or even the ULF tree, enabling easier inference. Second, the constraint solver is applied to the query frame. It resolves the main predicate and its arguments and determines which combinations of objects present in the scene satisfy the constraints contained in the question. The constraint solver returns these combinations along with the certainty for each combination as the answer. Based on this answer set, an answer to the user's question is generated in plain English. The response generator provides expanded answers for the given questions or reports failure in case the question constraints cannot be satisfied. The English answer is then sent to the web-app controlling the avatar for vocalizing it to the user, whereupon the system is ready for the next round.

\subsection{Unscoped Logical Form (ULF)} \label{sect:ulf}
We rely on ULF \cite{kim-schubert-2019-type} as an intermediate question representation format. ULF is similar in purpose to the {\it abstract meaning representation} (AMR) in semantic parsing~\cite{banarescu2013abstract}. However, ULF is close to the surface form of English, and covers a richer set of semantic phenomena than AMR,  and does so in a type-consistent way. To illustrate the approach, consider the example ``Which blocks are on two other blocks?''. The resulting ULF will be (((Which.d (plur block.n)) ((pres be.v) (on.p (two.d (other.a (plur block.n)))))) ?). As can be seen from this example, the resulting ULF retains much of the surface structure, but uses semantic typing and adds operators to indicate plurality, tense, aspect, and other linguistic phenomena. To facilitate conversion of English into ULF, a limited semantic parser was written and incorporated into the dialog manager, based on the typical vocabulary and phrasings occurring in spatial questions.

\subsection{Query Frames}
The query frame is the final representation of the question with all the components assigned to the designated slots. The structure of a query frame is recursive. An example is presented below:

\begin{itemize}[noitemsep]
    \item[] Sentence \{ 
    \item[] \hspace{3mm} Content = Predicate \{
    \item[] \hspace{12mm} Content = TPrep \{on\}
    \item[] \hspace{12mm} ARG0 = Argument \{ 
    \item[] \hspace{15mm} ObjectType = block,
    \item[] \hspace{15mm} ObjectId = NULL,
    \item[] \hspace{15mm} Determiner =  which,
    \item[] \hspace{15mm} Modifiers = [plur] \}
    \item[] \hspace{12mm} ARG1 = Argument \{ 
    \item[] \hspace{15mm} ObjectType = block,
    \item[] \hspace{15mm} ObjectId = NULL,
    \item[] \hspace{15mm} Determiner =  other,
    \item[] \hspace{15mm} Modifiers = [plur, TNumber \{two\}] \}
    \item[] \hspace{12mm} PredModifiers = [] \}\}
\end{itemize}

The \textit{Sentence} frame encapsulates all the components. Its top-level entries consist of \textit{Predicate} frames and \textit{Argument} frames, which encapsulate the spatial relations and the entities in the blocks world, respectively. The predicate frames contain the predicate itself, its arguments, and possible predicate modifiers, e.g., \textit{directly}, \textit{slightly}, etc. The argument frames contain the type of the argument (block, table, structure, or a region -- e.g., \textit{front of the table}). The \textit{ObjectId} slot is reserved for potential unique identifiers such as block names, and, potentially, unique structure identifiers. The \textit{Determiner} slot is of course reserved for determiners, and the \textit{Modifiers} slot contains a list of possible argument modifiers, which can be block colors, numerals, etc., perhaps including other predicates.

\subsection{Constraint Solver}
The constraint solver takes the above query frame as an input and processes it recursively. It should be noted here that the term ``constraint solver'' in our work does not refer to the general constraint satisfaction area, but designates a custom algorithm that resolves the arguments of the relations and finds the sets of objects satisfying the verbal constraints contained in the question. If the sentence contains a descriptive predicate, it first resolves its arguments, by taking the set of all the entities present in the context and applying a series of filters, so as to extract just the objects satisfying the specified attributes of the arguments. Then it resolves the predicate by mapping the content of the predicate (a word) to the actual function implementing that predicate. The constraint solver applies the resolved function to all combinations of arguments and generates a list of tuples of form ((\textit{arg0}, \textit{arg1}), \textit{certainty}) where \textit{certainty} is the numerical value representing how strongly the relation holds between the arguments. For example, for the question \textit{``Which blue blocks are on top of the Toyota block and the Burger King Block''}, the argument resolution will return the set of blue blocks as \textit{arg0}, e.g., (\textit{Target}, 1.0), (\textit{Texaco}, 1.0), (\textit{Toyota}, 1.0) (assuming that the \textit{Target}, \textit{Texaco} and the \textit{Toyota} blocks are blue), and a pair ((\textit{Toyota}, \textit{Burger King}), 1.0) as \textit{arg1}. In this example, since the color and identity of each block is known, the certainties are 1.0. Then the constraint solver will apply the predicate \textit{on} to each combination, obtaining a list like ((\textit{Target}, (\textit{Toyota}, \textit{Burger King})), 0.87), ((\textit{Texaco}, (\textit{Toyota}, \textit{Burger King})), 0.76), ((\textit{Toyota}, (\textit{Toyota}, \textit{Burger King})), 0.0), where from the certainty values we can infer that the \textit{Target} block is on top of both the \textit{Toyota} and the \textit{Burger King} block, the \textit{Texaco} block is somewhat on top, and the \textit{Toyota} block is definitely not.

The iterative filtering for the argument resolution starts with the set of objects present in the current active context. Then the objects are filtered according to their type, i.e., if the argument is a block with some additional modifiers, i.e., ``a red block", the constraint solver extracts all the objects that are blocks. The resulting set forms the initial candidate argument set. The constraint solver then goes through the modifiers, which can be articles, numerals, adjectives, or even nested spatial relations, and consecutively applyies the corresponding filters to the current candidate set to (usually) narrow it down.

The spatial relations are processed in a somewhat similar way; once the arguments are resolved and the predicate function has been applied to the set of argument tuples, that set is passed through a set of filters for relation modifiers, such as ``directly'', ``fully'', ``not'', etc.

\subsection{Spatial Relations}
Our work uses a rule-based approach and imagistic scene representation for computing spatial relations. Each spatial relation is a probabilistic predicate, taking one or more arguments and applying a sequence of metrics and rule checks, such as the distance between the object centers, whether the objects are in contact, whether they possess certain properties, etc. Each metric returns a real number from the interval $[0, 1]$. These metrics represent contributing factors to a relation and they are either linearly combined, or the maximum among them is taken, depending on the relation. Whenever possible we rely on approximations to the real 3D meshes of objects, using centroids and bounding boxes (smallest rectangular regions encompassing the objects). There are two main reasons for that. First, we are trying to achieve real-time performance. Second, in many circumstances, given the object shapes and distances between them, the approximations yield acceptable results. Among the basic geometric primitives used in our models are various scaled distances and vector relations. 

For example, for the ``in front of'' relation, we consider two cases. First, the so called \textit{deictic} ``in front of'' is the version based on the observer's coordinate system. We consider one object to be in front of the other, if it is closer to us, and its image on the retina overlaps with, or is very close to the image of the second object. We compute approximate 2D projections of the objects to the observer's visual plane (observer's position is modeled in Blender) and then compare the distances from the objects' centers and the observer and between the projections' centers to estimate the value for the deictic version of the relation. The second version, the \textit{extrinsic} ``in front of'' is based on the global coordinate system, that is, the inherent coordinate system of the table and the blocks. Even if an object $A$ is not in front of another object $B$ visually, it can still be considered to be in front of $B$, if it is inside a cone area originating at $B$ with the cone expanding towards the front of the table. After computation of both versions of the relation, the maximum of the two values is returned and the system concludes whether the ``in front of'' relation holds based on that.

The factors that contribute to the semantics of spatial prepositions can be divided into geometric and non-geometric (functional). Geometric factors are relatively straightforward; they include locations, sizes and distances. Non-geometric factors include background knowledge about the relata---their physical properties, roles, the way we interact with them---as well as the perceived ``frame" (such as a table top) and the presence and characteristics of other objects within that frame. Because of the simplicity and uniformity of objects and structures in the blocks world, geometric factors are the most important for our system. 
%We rely on 3D model of the scene with all the objects represented as polygonal meshes, so the geometric components of the prepositions can be directly inferred from the coordinates of points comprising the object's model. 
%We add additional geometric and non-geometric knowledge about the objects by manually attaching labels or tags to the meshes. 
%scaled by object dimensions,
An example is the distance between object centroids divided by the object sizes; that quantity will be 1.0 for cubes or spheres touching one another.
%, to the a measure relative to scale of the configuration. 
%Given two ideally-shaped objects (cubes or spheres) the scaled distance between them will be equal to 1 exactly when they are touching each other.
This is a useful measure if the objects are convex or located relatively far apart. We introduce separate distance metrics for %certain types of objects that are not compact, i.e., poorly approximated by a sphere, for example, 
flat (roughly planar) or elongated (roughly linear) objects such as the table, or stacks and rows
%the table itself and various structures such as stacks and rows, which can be approximated by planes and lines,
respectively.

%Another important geometric primitive is an infinite conic region, defined at a vertex by an orientation vector and the angular width of the cone. This primitive is used in computing so-called projective prepositions, such as \textit{above}, etc. For prepositions like \textit{to the left/right of}, whose value depends on the observer's vantage point, we project the arguments' meshes onto the observer's visual plane (orthogonal to its frontal or ``view'' vector) and then work with 2D data, either bounding boxes or entire mesh projections.

The perceived frame, and the scale and statistics of objects in the vicinity of objects being related, are additional important factors. For some prepositions we first compute the raw value (between 0 and 1) representing the context-independent value of the preposition's metric. This metric is then modified by scaling it up or down depending on the values of the same metric for other objects in the scene. For example, suppose that the raw nearness metric $near\_raw(A, B)$ for two objects $A$ and $B$ is 0.55 out of 1.0. This reflects the fact that without further context, this is an ambiguous situation. However, if $B$ is the closest object to $A$, i.e., $near\_raw(C, A) < 0.55, \forall C(C \not= B)$, we can say that $B$ is \textit{relatively} near $A$. In this case the final score $near(A, B)$ will be boosted by a small amount (depending on the distribution of the objects in the scene), making a more definite judgment possible.

The ``where is...''  questions are handled differently. For those, given the block, the system checks which combination of the relation and the second argument gives the highest value and returns them as the answer.

\subsection{Dialogue Manager} \label{sect:eta}
Eta is a dialogue manager (DM) designed to follow a modifiable dialogue schema, specified using a flexible and expressive schema language. This schema specifies a plan of expected dialogue actions of the user and system, subject to change as the interaction proceeds. The DM resembles the dialogue manager used by the LISSA system \cite{razavi2016IVA,razavi2017ACS}, but allows for logical interpretation of queries. The dialogue actions are instantiated into events over the course of the conversation. Possible actions include ``primitive'' explicitly-defined utterances by the system, as well as abstract actions to interpret the user's response, or to form a reaction (making a reply, or initiating a subplan) to this interpretation. Using hierarchical pattern transduction methods, these abstract actions are converted into one or more primitive actions during the execution of the dialogue plan. The pattern transduction process is implemented using transduction trees, whose internal nodes specify patterns to be matched against the expression to be transduced. When all patterns at the internal nodes of a root-to-leaf path have been matched, a template associated with the final node (leaf) is filled in using match results from the parent. If a match at an internal node fails, its sibling nodes are tried, and if all siblings fail, the search continues recursively at the siblings of the parent. The result of a tree transduction, i.e., a filled-in template, is used in whatever way is specified by a directive associated with the template (send the result to output, continue in some subtree, initiate a subschema, etc.).

The system uses hierarchical pattern transductions to interpret the user input in two stages. First, the DM extracts a simple, context-independent {\it gist clause} by ``tidying up'' the user's response in the context of Eta's previous utterance. Second, if the system recognizes the gist clause as a spatial question, the gist clause is transduced into its corresponding ULF representation. This latter transduction amounts to a semantic parse using transduction trees geared towards various phrase types (NPs, PPs VPs, etc.); templates at leaf nodes in this case specify assembly of the ULFs of subphrases, obtained recursively, into a complete ULF for the targeted type of phrase. The resulting question ULF is then output to the spatial question-answering system. To generate nontechnical verbal reactions to the user, the system uses pattern transduction to construct an output from the gist clause extracted in the previous step. In the case where the gist clause represents a spatial question, the spatial question-answering system supplies the response.

The dialogue manager also uses a limited coreference module, which can resolve anaphora and referring expressions such as ``it'', ``that block'', etc. by detecting and storing discourse entities in context and employing recency and syntactic salience heuristics. A more general coreference module which ranks mentions based on a set of features and weights is currently being developed.

\subsection{Response Generation and Communicating Back to User}
The response generation for spatial questions directly maps outputs from the constraint solver -- in the form  $<$arg0, relation, arg1, certainty$>$ -- to natural English answers, phrased differently for each type of spatial question that may be posed. Each question class is handled distinctly, altering the English phrasing of 
%the natural English interpretation of 
the constraint solver's answer so that it sounds natural and informative. However, more than just the question type is accounted for in the response. 
%often even the particular phrasing of a question subtlety changes the generated response.  
For example, most questions reveal the questioner's expectations about the desired number of answers through use of singular or plural terms, e.g., {\it ``which block is...'' vs. ``which blocks are...''}. If the system finds only one object satisfying the constraints when the phrasing of the question presupposes multiple ones (such as in the latter case), the response includes corrective phrasing, such as {\it ``Only the Toyota block...''}.  Furthermore, responses typically reflect the system's degree of certainty, and in cases where the system gives both certain and uncertain answers, the response separates them and distinguishes the confident answers from the uncertain ones. When the surface form of the answer is ready, it is sent to the avatar app to vocalize it for the user.

\section{A First Evaluation}

% Ben: ULF statistics
%
% Total # of spatial questions: 635
% # interpreted correctly: 470
% # interpreted incorrectly: 165
% # incorrect due to ASR error: 87
%
% % accuracy: 74.02%
% % wrong due to ASR error: 52.73%
% 
% Common errors:
% 1. 'standard' ASR mistakes of words, e.g. "Starbucks block" to "Starbucks bloke" - these can be added to preprocessing, but new ones frequently come up
% 2. ASR mistakes of prepositions/connectives are somewhat more problematic, i.e. "between the Starbucks block in the SRI block"
% 3. ASR cutting off the middle of a query
% 4. Not all prepositions are supposed in the parser, i.e. "what blocks are by the NVidia block?"
% 5. Some forms of superlatives are not supported, like "what block is the highest?", "what block is above all the others?".
% 6. Some conjoined propositions aren't supported, e.g. "nearest to", "closest to", "furthest from".
% 7. Indexical questions, e.g. "which block did I just move", "what blocks can you see", etc.
% 8. Some bugs with negation in queries
% 9. Passive expressions, e.g. "how many blocks are being touched by ..."

Since we are continuing to refine the system, our experimental assessment remains of modest scope. We enlisted the help of 5 volunteers in our department to test the capability of the system. Among the participants were both graduate and undergraduate students, both native and non-native English speakers. The participants were instructed to ask spatial questions of the general type supported by the system, but without restriction on wording; before their first session they were shown a short demonstration of the expected kind of interaction with the system, including a few question-answer exchanges. During each evaluation session they were requested to ask between 40 and 50 questions and mark the system's answer as correct, partially correct or incorrect. Additionally, if no answer could be given because of speech recognition errors, they were asked to indicate that as well. Finally, they were asked to indicate if the answer of the system (regardless of correctness) seemed to be improperly or oddly phrased. Each session started with the blocks positioned in a row at the front of the table. The participants were instructed to move the blocks arbitrarily to test the robustness and consistency of the spatial models. The data are presented in Table \ref{evaltab}.

\begin{table}[ht]
\caption{Evaluation data.}\smallskip
\centering
\resizebox{.95\columnwidth}{!}{
\smallskip\begin{tabular}{|c|c|}
    \hline
    Total number of questions & 388 \\
    \hline
    Correct answers & 219 \\
    \hline
    Partially correct answers & 45 \\
    \hline 
    Incorrect answers & 65 \\
    \hline 
    Errors in speech recognition & 59 \\
    \hline
    The answer was given but sounded unnatural/ungrammatical & 25 \\
    \hline
\end{tabular}
}
\label{evaltab}
\end{table}

The accuracy of the system for the correctly parsed questions is given in the Table \ref{acctab}.

\begin{table}[ht]
\caption{Accuracy for the parsed questions.}\smallskip
\centering
\resizebox{.85\columnwidth}{!}{
\smallskip\begin{tabular}{|c|c|}
    \hline
    Correct answers & 219 out of 329 (66.6\%) \\
    \hline
    Partially correct answers & 45 out of 329 (13.7\%) \\
    \hline 
    Incorrect answers & 65 out of 329 (19.8\%) \\
    \hline
\end{tabular}
}
\label{acctab}
\end{table}

We have found that the system is capable of correctly answering the great majority of spatial questions in the development set, and around 67\% of miscellaneous ``off the cuff" spoken questions asked by the participants during live dialogue test runs. Including partially correct ones, the accuracy rise to 80\%. Correctness is tracked both in terms of the ULFs produced (and displayed under the ``David" avatar) and in terms of the generated spoken answers. The spatial component also displays satisfactory sensitivity in terms of the certainty cut-off threshold. That is, the threshold determining which objects are included seems in accord with human intuitions. Correctness data obtained for the current version, while not very high in absolute terms, is actually quite satisfactory. After all, the semantics of spatial relations are vague and different people have slightly different ideas of what can be counted as, e.g., ``to the left of''. In fact, in our previously cited study we had observed an interannotator agreement score of only 0.72 on a 5-point Likert scale for a set of prepositional relations in the simulated room and blocks worlds \cite{platonov2018computational}.

Below we present separate evaluation data for the ULF parser.

\begin{table}[ht]
\caption{Evaluation data on ULF parsing.}\smallskip
\centering
\resizebox{.95\columnwidth}{!}{
\smallskip\begin{tabular}{|c|c|}
    \hline
    Total number of spatial questions & 635 \\
    \hline
    Number of correctly interpreted questions & 470 \\
    \hline
    Number of incorrectly interpreted questions & 165 \\
    \hline
    Number of incorrect parses due to ASR errors & 87 \\
    \hline
    Accuracy & 74.02\% \\
    \hline
    Percentage of incorrect parses due to ASR errors & 52.73\% \\
    \hline
\end{tabular}
}
\label{ulftab}
\end{table}

Errors in the ULF parsing fall into a few general categories:

\begin{itemize}
    \item ASR errors: This includes relatively straightforward word errors, such as outputting ``Starbucks bloke'' instead of ``Starbucks block''. However, the ASR can also make errors on important prepositions and connectives, which is somewhat more problematic. For example, ``between the Starbucks block in the SRI block'' (instead of ``and''). Finally, sometimes the ASR cuts off in the middle of a sentence.
    \item Unsupported sentence constructions: This includes some prepositions (particularly multiple-word items), e.g. ``Which blocks are by the NVidia block?'', ``what block is above all the others?'', etc. Passive expressions, e.g. ``How many blocks are touched by ...'' are also currently unsupported, and were used rarely in the dataset. Finally, there are a few bugs with parsing negations in queries.
    \item Indexical questions: On occasion users in the experiments would ask indexical questions which require interaction with context, such as ``What block did I just move?'', ``What blocks can you see?'', etc. The dialogue manager currently does not have the ability to interact with dialogue context in this way, however this component is being actively worked on.
\end{itemize}

% Common sources of errors include
% \begin{itemize}
%     \item Common ASR errors, e.g., "Starbucks bloke" instead of "Starbucks block"
%     \item ASR mistakes of prepositions/connectives are somewhat more problematic, i.e. "between the Starbucks block in the SRI block"
%     \item ASR cutting off the middle of a query
%     \item Not all prepositions are supposed in the parser, i.e. "what blocks are by the NVidia block?"
%     \item Some forms of superlatives are not supported, like "what block is the highest?", "what block is above all the others?".
%     \item Some conjoined propositions aren't supported, e.g. "nearest to", "closest to", "furthest from".
%     \item Indexical questions, e.g. "which block did I just move", "what blocks can you see", etc.
%     \item Some bugs with negation in queries
%     \item Passive expressions, e.g. "how many blocks are being touched by ..."
% \end{itemize}

Some commonly occurring speech recognition mistakes (many of them are caused by block names) were taken into account and fixed during the preprocessing stage, however some of the errors occur rarely due to a dependence of a particular speaker accent and/or expression being uttered, and are hard to catch.

\section{Conclusion}
We have built a spatial question answering system for a physical blocks world, already able to handle a majority of questions in dialogue mode. We are not aware of any other end-to-end system with comparable abilities in QA dialogues about spatial relations. Our spatial language model relies on intuitive computational models of spatial prepositions that come close to mirroring human judgments by combining geometrical information with context-specific information about the objects and the scene. This enables natural interaction between the machine and the user. The ongoing work and near-term work is targeting reasoning about structures and complex shapes, which will eventually be incorporated into blocks world structure learning and collaborative construction tasks.

\section{Acknowledgments}
%Acknowledgements are omitted for anonymity.
This work was supported by DARPA grant W911NF-15-1-0542. We thank our team of volunteers for their suggestions and contributions to system evaluation.

\bibliographystyle{aaai}
\bibliography{refs}

\begin{thebibliography}{}

\bibitem[\protect\citeauthoryear{Banarescu \bgroup et al\mbox.\egroup
  }{2013}]{banarescu2013abstract}
Banarescu, L.; Bonial, C.; Cai, S.; Georgescu, M.; Griffitt, K.; Hermjakob, U.;
  Knight, K.; Koehn, P.; Palmer, M.; and Schneider, N.
\newblock 2013.
\newblock Abstract meaning representation for sembanking.
\newblock In {\em Proceedings of the 7th Linguistic Annotation Workshop and
  Interoperability with Discourse},  178--186.

\bibitem[\protect\citeauthoryear{Bigelow \bgroup et al\mbox.\egroup
  }{2015}]{bigelow2015need}
Bigelow, E.; Scarafoni, D.; Schubert, L.; and Wilson, A.
\newblock 2015.
\newblock On the need for imagistic modeling in story understanding.
\newblock {\em Biologically Inspired Cognitive Architectures} 11:22--28.

\bibitem[\protect\citeauthoryear{Bisk \bgroup et al\mbox.\egroup
  }{2018}]{bisk2018learning}
Bisk, Y.; Shih, K.~J.; Choi, Y.; and Marcu, D.
\newblock 2018.
\newblock Learning interpretable spatial operations in a rich 3d blocks world.
\newblock In {\em Thirty-Second AAAI Conference on Artificial Intelligence}.

\bibitem[\protect\citeauthoryear{Boteanu \bgroup et al\mbox.\egroup
  }{2016}]{boteanu2016model}
Boteanu, A.; Howard, T.; Arkin, J.; and Kress-Gazit, H.
\newblock 2016.
\newblock A model for verifiable grounding and execution of complex natural
  language instructions.
\newblock In {\em Intelligent Robots and Systems (IROS), 2016 IEEE/RSJ
  International Conference on},  2649--2654.
\newblock IEEE.

\bibitem[\protect\citeauthoryear{Broad \bgroup et al\mbox.\egroup
  }{2016}]{broad2016towards}
Broad, A.; Arkin, J.; Ratliff, N.; Howard, T.; Argall, B.; and Graph, D.~C.
\newblock 2016.
\newblock Towards real-time natural language corrections for assistive robots.
\newblock In {\em RSS Workshop on Model Learning for Human-Robot
  Communication}.

\bibitem[\protect\citeauthoryear{Chang, Savva, and
  Manning}{2014}]{chang2014learning}
Chang, A.; Savva, M.; and Manning, C.~D.
\newblock 2014.
\newblock Learning spatial knowledge for text to 3d scene generation.
\newblock In {\em Proceedings of the 2014 Conference on Empirical Methods in
  Natural Language Processing (EMNLP)},  2028--2038.

\bibitem[\protect\citeauthoryear{Chen \bgroup et al\mbox.\egroup
  }{2015}]{chen2015survey}
Chen, J.; Cohn, A.~G.; Liu, D.; Wang, S.; Ouyang, J.; and Yu, Q.
\newblock 2015.
\newblock A survey of qualitative spatial representations.
\newblock {\em The Knowledge Engineering Review} 30(01):106--136.

\bibitem[\protect\citeauthoryear{Chung \bgroup et al\mbox.\egroup
  }{2015}]{chung2015performance}
Chung, I.; Propp, O.; Walter, M.~R.; and Howard, T.~M.
\newblock 2015.
\newblock On the performance of hierarchical distributed correspondence graphs
  for efficient symbol grounding of robot instructions.
\newblock In {\em Intelligent Robots and Systems (IROS), 2015 IEEE/RSJ
  International Conference on},  5247--5252.
\newblock IEEE.

\bibitem[\protect\citeauthoryear{Cohn and Renz}{2008}]{cohn2008qualitative}
Cohn, A.~G., and Renz, J.
\newblock 2008.
\newblock Qualitative spatial representation and reasoning.
\newblock {\em Foundations of Artificial Intelligence}  551.

\bibitem[\protect\citeauthoryear{Collell, Van~Gool, and
  Moens}{2017}]{collell2017acquiring}
Collell, G.; Van~Gool, L.; and Moens, M.-F.
\newblock 2017.
\newblock Acquiring common sense spatial knowledge through implicit spatial
  templates.
\newblock {\em arXiv preprint arXiv:1711.06821}.

\bibitem[\protect\citeauthoryear{Dhelim, Ning, and Zhu}{2016}]{dhelim2016stlf}
Dhelim, S.; Ning, H.; and Zhu, T.
\newblock 2016.
\newblock Stlf: Spatial-temporal-logical knowledge representation and object
  mapping framework.
\newblock In {\em Systems, Man, and Cybernetics (SMC), 2016 IEEE International
  Conference on},  001550--001554.
\newblock IEEE.

\bibitem[\protect\citeauthoryear{Fahlman}{1974}]{fahlman1974planning}
Fahlman, S.~E.
\newblock 1974.
\newblock A planning system for robot construction tasks.
\newblock {\em Artificial intelligence} 5(1):1--49.

\bibitem[\protect\citeauthoryear{Fikes and Nilsson}{1971}]{fikes1971strips}
Fikes, R.~E., and Nilsson, N.~J.
\newblock 1971.
\newblock {STRIPS}: A new approach to the application of theorem proving to
  problem solving.
\newblock {\em Artificial Intelligence} 2(3-4):189--208.

\bibitem[\protect\citeauthoryear{Howard, Tellex, and
  Roy}{2014}]{howard2014natural}
Howard, T.~M.; Tellex, S.; and Roy, N.
\newblock 2014.
\newblock A natural language planner interface for mobile manipulators.
\newblock In {\em Robotics and Automation (ICRA), 2014 IEEE International
  Conference on},  6652--6659.
\newblock IEEE.

\bibitem[\protect\citeauthoryear{Johnson \bgroup et al\mbox.\egroup
  }{2017}]{johnson2017clevr}
Johnson, J.; Hariharan, B.; van~der Maaten, L.; Fei-Fei, L.; Lawrence~Zitnick,
  C.; and Girshick, R.
\newblock 2017.
\newblock Clevr: A diagnostic dataset for compositional language and elementary
  visual reasoning.
\newblock In {\em Proceedings of the IEEE Conference on Computer Vision and
  Pattern Recognition},  2901--2910.

\bibitem[\protect\citeauthoryear{Kim and
  Schubert}{2019}]{kim-schubert-2019-type}
Kim, G.~L., and Schubert, L.
\newblock 2019.
\newblock A type-coherent, expressive representation as an initial step to
  language understanding.
\newblock In {\em Proceedings of the 13th International Conference on
  Computational Semantics - Long Papers},  13--30.
\newblock Gothenburg, Sweden: Association for Computational Linguistics.

\bibitem[\protect\citeauthoryear{Kottur \bgroup et al\mbox.\egroup
  }{2019}]{kottur2019clevr}
Kottur, S.; Moura, J.~M.; Parikh, D.; Batra, D.; and Rohrbach, M.
\newblock 2019.
\newblock Clevr-dialog: A diagnostic dataset for multi-round reasoning in
  visual dialog.
\newblock {\em arXiv preprint arXiv:1903.03166}.

\bibitem[\protect\citeauthoryear{Liu \bgroup et al\mbox.\egroup
  }{2019}]{liu2019clevr}
Liu, R.; Liu, C.; Bai, Y.; and Yuille, A.~L.
\newblock 2019.
\newblock Clevr-ref+: Diagnosing visual reasoning with referring expressions.
\newblock In {\em Proceedings of the IEEE Conference on Computer Vision and
  Pattern Recognition},  4185--4194.

\bibitem[\protect\citeauthoryear{Mao \bgroup et al\mbox.\egroup
  }{2019}]{mao2019neuro}
Mao, J.; Gan, C.; Kohli, P.; Tenenbaum, J.~B.; and Wu, J.
\newblock 2019.
\newblock The neuro-symbolic concept learner: Interpreting scenes, words, and
  sentences from natural supervision.
\newblock {\em arXiv preprint arXiv:1904.12584}.

\bibitem[\protect\citeauthoryear{Paul \bgroup et al\mbox.\egroup
  }{2016}]{paul2016efficient}
Paul, R.; Arkin, J.; Roy, N.; and Howard, T.
\newblock 2016.
\newblock Efficient grounding of abstract spatial concepts for natural language
  interaction with robot manipulators.
\newblock {\em Proceedings of Robotics: Science and Systems (RSS), Ann Arbor,
  Michigan, USA}.

\bibitem[\protect\citeauthoryear{Paul \bgroup et al\mbox.\egroup
  }{2018}]{paul2018efficient}
Paul, R.; Arkin, J.; Aksaray, D.; Roy, N.; and Howard, T.~M.
\newblock 2018.
\newblock Efficient grounding of abstract spatial concepts for natural language
  interaction with robot platforms.
\newblock {\em The International Journal of Robotics Research}
  0278364918777627.

\bibitem[\protect\citeauthoryear{Perera \bgroup et al\mbox.\egroup
  }{2018a}]{perera2018building}
Perera, I.; Allen, J.; Teng, C.~M.; and Galescu, L.
\newblock 2018a.
\newblock Building and learning structures in a situated blocks world through
  deep language understanding.
\newblock In {\em Proceedings of the First International Workshop on Spatial
  Language Understanding},  12--20.

\bibitem[\protect\citeauthoryear{Perera \bgroup et al\mbox.\egroup
  }{2018b}]{perera2018situated}
Perera, I.; Allen, J.; Teng, C.~M.; and Galescu, L.
\newblock 2018b.
\newblock A situated dialogue system for learning structural concepts in blocks
  world.
\newblock In {\em Proceedings of the 19th Annual SIGdial Meeting on Discourse
  and Dialogue},  89--98.

\bibitem[\protect\citeauthoryear{Platonov and
  Schubert}{2018}]{platonov2018computational}
Platonov, G., and Schubert, L.
\newblock 2018.
\newblock Computational models for spatial prepositions.
\newblock In {\em Proceedings of the First International Workshop on Spatial
  Language Understanding},  21--30.

\bibitem[\protect\citeauthoryear{Razavi \bgroup et al\mbox.\egroup
  }{2016}]{razavi2016IVA}
Razavi, S.; Ali, M.; Smith, T.; Schubert, L.; and Hoque, M.
\newblock 2016.
\newblock The {LISSA} virtual human and {ASD} teens: An overview of initial
  experiments.
\newblock In {\em Proc. of the 16th Int. Conf. on Intelligent Virtual Agents
  (IVA 2016)},  460--463.

\bibitem[\protect\citeauthoryear{Razavi \bgroup et al\mbox.\egroup
  }{2017}]{razavi2017ACS}
Razavi, S.; Schubert, L.; Ali, M.; and Hoque, H.
\newblock 2017.
\newblock Managing casual spoken dialogue using flexible schemas, pattern
  transduction trees, and gist clauses.
\newblock In {\em 5th Ann. Conf. on Advances in Cognitive Systems (ACS 2017)}.

\bibitem[\protect\citeauthoryear{Regier}{1996}]{regier1996human}
Regier, T.
\newblock 1996.
\newblock {\em The human semantic potential: Spatial language and constrained
  connectionism}.
\newblock MIT Press.

\bibitem[\protect\citeauthoryear{Roy, Hsiao, and
  Mavridis}{2004}]{roy2004mental}
Roy, D.; Hsiao, K.-Y.; and Mavridis, N.
\newblock 2004.
\newblock Mental imagery for a conversational robot.
\newblock {\em Systems, Man, and Cybernetics, Part B: Cybernetics, IEEE
  Transactions on} 34(3):1374--1383.

\bibitem[\protect\citeauthoryear{Tellex \bgroup et al\mbox.\egroup
  }{2011}]{tellex2011understanding}
Tellex, S.; Kollar, T.; Dickerson, S.; Walter, M.~R.; Banerjee, A.~G.; Teller,
  S.; and Roy, N.
\newblock 2011.
\newblock Understanding natural language commands for robotic navigation and
  mobile manipulation.
\newblock In {\em Twenty-Fifth AAAI Conference on Artificial Intelligence}.

\bibitem[\protect\citeauthoryear{Winograd}{1972}]{winograd1972understanding}
Winograd, T.
\newblock 1972.
\newblock Understanding natural language.
\newblock {\em Cognitive psychology} 3(1):1--191.

\bibitem[\protect\citeauthoryear{Yu and Siskind}{2017}]{yu2017sentence}
Yu, H., and Siskind, J.~M.
\newblock 2017.
\newblock Sentence directed video object codiscovery.
\newblock {\em International Journal of Computer Vision} 124(3):312--334.

\end{thebibliography}
\end{document}